%% file: main.tex
\documentclass[letterpaper, 10 pt, conference]{ieeeconf}
\IEEEoverridecommandlockouts                              

\usepackage[utf8]{inputenc}
\usepackage{graphicx}
\usepackage{algorithm2e}
\usepackage{xcolor}
\usepackage{amsmath}
\usepackage{amssymb}

\usepackage{multicol}

\title{\LARGE \bf 
An Optimization-Based Inverse Kinematics Solver for Continuum Manipulators in Intricate Environments}
\author{Yinan Sun, Sai Wang
}

\begin{document}

\maketitle

\begin{abstract}
Continuum manipulators have gained significant attention as a promising alternative to rigid manipulators, offering notable advantages in terms of flexibility and adaptability within intricate workspace. However, the broader application of high degree-of-freedom (DoF) continuum manipulators in intricate environments with multiple obstacles necessitates the development of an efficient inverse kinematics (IK) solver specifically tailored for such scenarios. Existing IK methods face challenges in terms of computational cost and solution guarantees for high DoF continuum manipulators, particularly within intricate workspace that obstacle avoidance is needed. To address these challenges, we have developed a novel IK solver for continuum manipulators that incorporates obstacle avoidance and other constraints like length, orientation, etc., in intricate environments, drawing inspiration from optimization-based path planning methods. 
Through simulations, our proposed method showcases superior flexibility, efficiency with increasing DoF, and robust performance within highly unstructured workspace, achieved with acceptable latency.

\end{abstract}
\section{Introduction}
Continuum manipulators \cite{singh2014continuum}, characterized by their continuous bending along the entire length, pose unique challenges for inverse kinematics. Traditional methods, such as Jacobian-based approaches \cite{rucker2011computing}, incur significant computational costs when applied to continuum manipulators due to their highly redundant degrees of freedom and continuous structure, which lack closed-form solutions \cite{lai2019learning}. However, these manipulators are prized for their compliance, safety, and capacity to navigate intricate, constrained workspaces \cite{santoso2021origami}.

In addressing the IK problem for continuum manipulators, a common assumption is that each finite section of the continuum body follows a constant curvature arc \cite{garriga2019kinematics, wen2023modeling}, allowing for a kinematic representation based on bending angle and arc length. The assumption is based on the fact that, most of the continuum manipulators are built by serially connect several bending actuators. These manipulators, with their high redundancy, often entail significant computational costs when applying conventional methods. 

Existing approaches to IK for continuum manipulators often require substantial computational resources and offer no guarantee of a correct solution. For instance, the use of Geometric Jacobian \cite{lin2023piecewise}, a prevalent choice, is encumbered by substantial computational demands, offering no assurance of a correct solution. Researchers have explored alternative approaches, such as Monte Carlo algorithms \cite{qin2022snake}, which, while effective, suffer from prolonged computational times, particularly in systems with numerous DoF. Searching-based algorithms, while promising, still incur substantial computational costs as the number of sections increases. Also, it is mainly used in manipulators with rigid links \cite{sinha2019geometric}. 

FABRIKc and its variants, including FABRIKx, are iterative algorithms designed to solve the IK problem for continuum manipulators \cite{zhang2018fabrikc,kolpashchikov2022fabrikx}. Inspired by the original FABRIK algorithm \cite{aristidou2011fabrik}, these methods iteratively adjust poses, incorporating both forward and backward iterations to achieve real-time solutions with smooth postures. However, these algorithms have limitations, such as multiple constraints accommodation and a lack of consideration for intricate workspaces with multiple obstacles. To address these limitations, Wu et al. \cite{wu2022novel} developed a heuristic algorithm based on FABRIK for obstacle avoidance. This algorithm performs collision checks and updates key points away from obstacles. However, it only activates when a collision check fails during iteration, which can lead to passive avoidance and potential proximity to dangerous areas near obstacles.

Learning-based algorithms have also been developed for solving the IK of continuum robots \cite{thuruthel2019soft, wu2021iterative}. This type of algorithm can be used in the model-free control of the continuum robot with computational efficiency, while the quality of the output relies largely on the training sets, which might lead to reducing robustness in obstacle settings that are way different from the training sets. Also, the training set would be too large for a redundant continuum robot.

Sampling-based algorithms, commonly used for path planning in rigid robotic arms, are also employed for solving inverse kinematics in continuum robots. For instance, researchers have combined rapidly-exploring random trees (RRT) and follow-the-leader (FTL) algorithms to address obstacle avoidance in continuum robots \cite{guochen2014path,niu2020path}. While these algorithms improve efficiency, they may not guarantee stable performance, especially in complex environments where obstacle settings require high accuracy for generating usable solutions considering obstacle avoidance.

Other algorithms also utilize path planning methods in IK solvers, such as SLINKI \cite{chiang2021slinki}, a hybrid IK solver that has made significant advancements in solving the IK problem for continuum manipulators. SLINKI utilizes state lattice-based A* search algorithms, a path planning method, to generate configurations for the initial $(n-2)$ segments with short latency. It then employs the FABRIKc algorithm for the accurate calculation of the remaining 2 segments, achieving a good balance between efficiency and accuracy. However, its time complexity may significantly increase with the number of segments considering the exponential computational complexity, posing challenges in computational efficiency as the DoF of the robot increases.

Despite the progress in IK for continuum robots, challenges remain. Existing methods may struggle with real-time performance as the number of DoFs increases, and they may not adequately consider multiple constraints or obstacle avoidance in complex environments.

In this paper, we present an optimization-based IK solver tailored for continuum manipulators, addressing IK challenges in intricate workspaces while accommodating constraints such as segment lengths and orientations. Our approach leverages the high degrees of freedom (DoFs) of continuum manipulators to navigate complex workspaces effectively.
Our methodology employs an optimization-based path planning algorithm with an artificial potential field to navigate multi-obstacle environments. The path is defined by path points, which are the endpoints of continuous piece-wise arcs. These arcs construct a continuous representation of the continuum robot. By calculating objective functions and employing optimization methods, we ensure that the continuum manipulator satisfies all constraints.

A key innovation of our approach is its ability to fully utilize the high DoF characteristic of continuum manipulators. We represent the manipulator as a holistic structure, making it easy to modify by adding more points on the path. Besides, our algorithm allows for obstacle avoidance even when the manipulator is not in direct contact with obstacles, leading to safer solutions.

Furthermore, our algorithm maintains computational efficiency, with a manageable time complexity of $O(n^2)$ as the number of segments increases. This efficiency ensures that our method remains practical and efficient for continuum manipulators of varying DoF.

The contributions of this research paper are outlined as follows:
\begin{itemize}
    \item We have developed an optimization-based framework for solving inverse kinematics (IK) for continuum robots, addressing various requirements such as end orientation tracking, length control, and intricate obstacle avoidance. This framework includes the formulation of a cost function and update rules essential for optimization.

    \item Our research potentially marks a pioneering achievement by showcasing that our IK solver enables the end effector of a continuum manipulator to track paths collision-free within intricate workspace replete with multiple obstacles. This capability sets the stage for enhanced manipulator performance and safety in challenging real-world scenarios.

    \item Our approach demonstrates the potential of efficient continuum manipulator IK by achieving an $O(n^2)$ complexity algorithm, offering practicality for manipulators with varying numbers of segments and high Degrees of Freedom.    
\end{itemize}

\section{Approach and Methods}

In addressing the challenges posed by the highly redundant structure of continuum manipulators, our methodology takes a holistic approach, treating the manipulator as a system of continuous paths. Central to our approach is the development of an innovative inverse kinematic solver designed to determine optimal configurations for each segment within the manipulator's structure.

Our solver focuses on creating a smooth path consisting of points that connect the two designated endpoints. These points serve as the endpoints of a piece-wise arc, which we later characterize by $G^1$ continuity.
Subsequently, we employ an optimization process to refine the initial configuration. This optimization considers length constraints associated with each segment, as well as obstacles, orientation, and other objective functions. This ensures the manipulator's operability within its environment.

This methodology forms the core of our research, enabling precise control and real-time operation of continuum manipulators, especially in intricate environments with complex obstacles.



\subsection{Piece-wise Arc Representation and Gradient Calculation}
\input{sections/arc_represent}

\subsection{Cost Function Build}
\label{sec:cost_function}
\input{sections/cost_functions}

\subsection{Update Rules}
\input{sections/update_rules}

\section{Examples and Results}
\input{sections/examples_n_results}

\section{Conclusion and Future Works}

In this research paper, we introduced a novel Inverse Kinematics solver designed specifically for continuum manipulators. Our approach integrates optimization-based path planning algorithms to establish an initial configuration and utilizes rational B\'ezier representation for arc length control. Through experiments, we demonstrate that the time consumption remains manageable even as the number of segments increases. Furthermore, we demonstrated the effectiveness of our IK solver in accurately tracking the end-effector path within complex environments featuring multiple obstacles.

For future work, this research paves the way for several avenues to enhance the efficiency and applicability of our algorithm. We intend to refine the updating mechanism by introducing adjustable waypoints and iteration numbers to optimize the IK solution further. Additionally, we will explore the integration of sampling-based path planning algorithms within the path generation process, with an array of more potential optimization objects. Extending our algorithm to three-dimensional scenarios represents another direction for future research.

Furthermore, we plan to implement our methodology on a physical manipulator equipped with touchless sensing techniques to enable real-time modeling of the surrounding environment. This will enable the development of an efficient motion planning framework tailored to continuum manipulators, capitalizing on the capabilities of our IK solver, to ensure rapid and practical applications in real-world scenarios.


\bibliographystyle{IEEEtran}
\bibliography{references}

\end{document}

%% file: sections/arc_represent.tex
To transition from path points to the continuum robot's configuration, we utilize a piece-wise arc representation that adheres to the constant curvature framework. Each segment of this piece-wise arc aligns with a corresponding segment in the continuum robot, ensuring $G^1$ continuity for coherence with adjacent robot segments.

Utilizing the piece-wise arc representation, we transform path points into a piece-wise arc connecting the path points and end points, aligning with the preset orientation at the start point. Subsequently, we optimize this parameterized piece-wise arc to align with constraints and avoid obstacles, employing an artificial potential field. This approach enables us to accurately represent the continuum manipulator's configuration while ensuring practical and functional configurations for real-world applications.

\subsubsection{Shape Parameters and Control Points}
\label{sec:arc_para}
To accurately represent the continuum manipulator's structure and achieve $G^1$ continuity, we employ a circular spline representation approach similar to Song et al.'s work \cite{song2009circular}. This approach involves using rational B'ezier curves to represent a single arc segment, with end points $a$ and $b$, and a control point $c$ lying on the bisector plane of the line segment $ab$. The arc equation using the rational B\'ezier representation is given by:

$$y(u)=\frac{(1-u)^2a\pm 2u(1-u)\omega c+u^2b}{(1-u)^2\pm 2u(1-u)\omega+u^2},u\in[0,1],$$ 
where the $\pm$ sign denotes the minor or major arc, and $\omega$ is determined by:

$$\omega =\frac{h}{\sqrt{h^2+k^2}}$$
Here, $h$ and $k$ are components of the vector perpendicular to the line segment $ab$.

For our piece-wise arc representation with $G^1$ continuity, we compute the control point $c$ of each arc segment based on the end points of the current arc segment and the control point $c$ of the previous segment.

\begin{figure}
    \centering
    \includegraphics[width=1.0\linewidth]{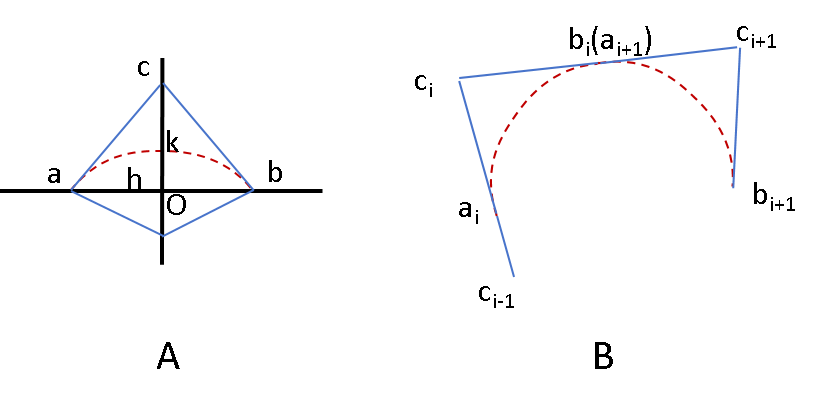}
    \caption{A) Rational B\'ezier representation of a single circular arc. B) A 2 segments piece-wise arc spline with $G^1$ continuity.
}      
    \label{arc}
\end{figure}

Figure~\ref{arc} shows the rational B\'ezier representation of a single circular arc, and a piece-wise arc spline with $G^1$ continuity.

For each arc segment, the control point $c_{i+1}$ is determined by the end points $a_{i+1}$ and $b_{i+1}$ of the segment, as well as the control point $c_{i}$ of the previous segment.

The control points are generated using the formula:
$$\vec{c_0a_0}=\vec{v_{iniori}}$$
Here, $\vec{v_{initial}}$ represents the preset orientation at the start point (or base) of the continuum robot. For subsequent control points, we use:
$$c_{i+1} = f(a_{i+1}, b_{i+1}, c_{i})$$,
where $f(a_{i+1}, b_{i+1}, c_{i})$ calculates the control point based on the end points and the previous control point. This approach allows us to create a piece-wise arc with path points and preset initial orientation, ensuring $G^1$ continuity and proper control point positioning.

To establish an initial guess for our optimization process, we create an arc connecting the start point and the goal (with the direction at the start point aligned with the initial orientation). We then divide this arc into a piece-wise arc with segments of equal length, treating the points on the arc as path points for the optimization process. With these equidistant points, we create the control points (the $c$ points) to ensure $G^1$ continuity within the piece-wise arc. This initial piece-wise arc serves as the starting configuration for our optimization process, providing a foundation for the subsequent optimization procedure.

\subsubsection{Gradient Calculation}
\label{sec:gradient_calc}
We consider the variation as the piece-wise arc itself, which is equivalent to optimizing the path points, as there is a bijection between the piece-wise arc and a set of path points with a fixed orientation at the start point. To effectively optimize the piece-wise arc, we calculate the gradient using numerical differentiation techniques with the following equation:
$$\frac{\partial F(p)}{p} = \frac{F(p+\delta p)-F(p-\delta p)}{2\delta p}$$
Here, $p$ represents the vector of path points' positions, $F(p)$ represents the objective function, and $\delta p$ represents a small disturbance on the path points.
While this method is straightforward and easy to deploy, it is crucial for managing potential issues of gradient explosion and dimension scaling. The disturbance of a path point can significantly affect the situation of an arc segment far away, similar to error accumulation.
To optimize the piece-wise arc effectively and mitigate the potential for gradient explosion, we use the following gradient descent equation:
$$dq_i = \alpha^{j-i}\sum_{j=i}^{n}\frac{\partial F_j}{\partial p_i}$$
Here, $dp_i$ represents the update value for the $i^{th}$ path point, and $\frac{\partial F_j}{\partial p_i}$ calculates the partial derivative representing the effect of the $i^{th}$ path point on the $j^th$ arc segment during optimization. The parameter $\alpha$ serves as a scaling factor crucial for managing dimension scaling and stabilizing the gradient descent process. It determines the rate at which previous points' contributions diminish and helps control the magnitude of parameter updates. Typically, $\alpha$ is set to a small value, such as $1e-2$, to ensure a controlled and stable optimization process.

All the symbols and explanations are listed in Table~\ref{table:arc_fitting_parameter}.

\begin{table}
    \caption{Parameters in the arc fitting and optimization.} 
	\centering 
	\begin{tabular}{l l} 
		\hline\hline 
		 Parameters & Explanation\\ [1.0ex]
		\hline 
     $a$& end point of an arc segment\\[1.0ex]
     $b$& another end point of an arc segment\\[1.0ex]
     $c$& control point of an arc segment, lies in the bisector\\&plane of line segment $ab$\\[1.0ex]
     $\omega$& weight parameter in the rational B\'ezier representation\\[1.0ex]
     $m$& central point of line segment $ab$\\[1.0ex]
     $n$& number of the continuum robot segments $ab$\\[1.0ex]
     $i$& index of arc segments\\[1.0ex]
     $h$& length of line segment $am$ and $mb$\\[1.0ex]
     $h$& length of line segment $cm$\\[1.0ex]
     $\alpha$& scale parameter preventing gradient exploding\\[1.0ex]
    
    \hline 
	\end{tabular}
	\label{table:arc_fitting_parameter} 
\end{table}

%% file: sections/cost_functions.tex
\subsubsection{Obstacle Avoidance}

The optimization-based algorithm utilizes a gradient descent mechanism that considers environmental obstacles and other constraints. This ensures that the path quickly converges to a locally optimal, collision-free, and continuous path, crucial for configuring a viable continuum manipulator. The approach integrates obstacle avoidance within the optimization framework, enhancing the manipulator's capability to navigate complex environments. The cost function ($F_{path}$) for obstacle avoidance is defined as follows:

$$F_{path} = \int_{0}^{1} c(\xi(t))||\frac{d}{dt}\xi(t)|| dt$$

Here, $\xi(t)$ represents the path consisting of path points, and $c(\xi)$ represents the cost with respect to obstacles, derived from an artificial potential field. The calculation of $c(\xi)$ is described by the following formula:

$$c(\xi(t))=\left\{
\begin{array}{cc}
     k_o(-d(\xi(t))+\frac{1}{2}\epsilon),& if~d(\xi(t))<0 \\
     \frac{k_o}{2\epsilon} (d(\xi(t))-\epsilon)^2,& if~0\leq d(\xi(t)) \leq \epsilon\\
     0, & otherwise
\end{array}
\right.
$$

\begin{figure}
    \centering
    \includegraphics[width=1.0\linewidth]{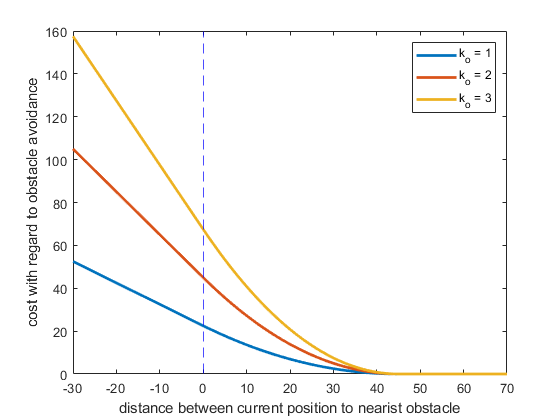}
    \caption{Cost function for obstacles with relate to distance between current position and the nearest obstacle.
}      
    \label{potential_field}
\end{figure}

$d(\xi)$ denotes the distance between the path $\xi$ and the nearest obstacle. $||\frac{d}{dt}\xi(t)||$ represents the 'velocity' of each path point. In this context, it refers to the mean value of the two arc segments adjacent to the current path point. This term is included in the cost function to discourage the behavior of enlarging the distance between path points to decrease the cost function value. $\epsilon$ signifies the maximum distance at which an obstacle still influences the path during the optimization procedure. The parameter $k_o$ is introduced to adjust the slope of the artificial potential field. A higher $k_o$ value leads to a steeper potential field, as illustrated in Figure~\ref{potential_field}.
The function $c(\xi(t))$ is designed to be $G^1$ continuous with respect to $d(\xi(t))$ to ensure robust convergence and minimize the likelihood of oscillations or other convergence issues during optimization. Additionally, the function is linearized when $d(\xi(t))<0$ to prevent excessively large gradients that could impact convergence.

To facilitate efficient computation of the cost function for obstacle avoidance, we precompute a discrete artificial potential field offline (in the real world, it can be aqcuired by remote sensing, Lidar, RGBD camera, etc.). This involves recording the cost function value at grid nodes in the workspace with a reasonable density. During the optimization process, we can efficiently calculate the cost function by interpolating the values at the nearest grid nodes, thereby avoiding the need for costly real-time calculations.

\begin{figure}
    \centering
    \includegraphics[width=0.8\linewidth]{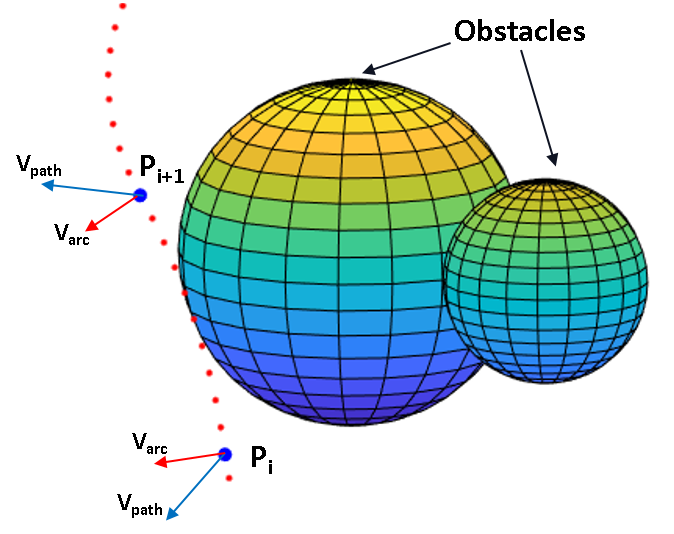}
    \caption{
    A sketch for obstacle avoidance. Blue points represent the path points, which are also the end points of each arc segment. Blue vectors are generated by the optimization-based path planning algorithm, aiming to guide the overall path towards lower potential positions in the obstacle-related artificial potential field. The red points are samples taken from the piece-wise arcs, representing the continuum robot body. The red vectors depict the gradients calculated using a numerical method to move the piece-wise arcs away from obstacles during optimization.
}      
    \label{arc_points}
\end{figure}
As a continuum robot, ensuring collision-free operation is not guaranteed solely by the segments' endpoints being collision-free. To address this, we utilize the rational B\'ezier representation equation mentioned in Section~\ref{sec:arc_para} to generate points along the continuum robot's body as a supplement, and for collision checking, as illustrated in Figure~\ref{arc_points}. Additionally, we employ the numerical gradient calculation equation from Section~\ref{sec:gradient_calc} to compute the gradient as part of the obstacle avoidance update vector. This allows the piece-wise arc to avoid obstacles, not just the path points. Consequently, the gradient related to obstacle avoidance is defined as:
$$\Delta F_{obs} = \Delta F_{path} + \beta\Delta F_{arcs}$$

Here, $\Delta F_{path}$ represents the update vector derived from $F_{path}$ to move the path points away from obstacles. $\Delta F_{arcs}$ represents the gradient calculated using the points sampled from the piece-wise arc, aiding the arc in avoiding obstacles. The parameter $\beta$ determines the weighting between these two gradients. $\Delta F_{obs}$ is the overall gradient for obstacle avoidance.

\subsubsection{Other Constraints }
To achieve precise control over the continuum manipulator's configuration, it is essential to manage the length of each segment within our inverse kinematics solver. Our methodology ensures that each segment closely aligns with the specified central length.

The optimization process begins by addressing the length of each segment within the piece-wise arc. Assuming the range of segment lengths is $[l_{\text{min}}, l_{\text{max}}]$, the objective function guiding this optimization is defined as:

$$F(p)=\sum_{i=1}^n (l_i-l_0)^2$$
Here, $l_i$ represents the actual length of the segment, and $l_0$ denotes the mean value of the range $l_{\text{max}}$ and $l_{\text{min}}$.

Subsequently, following the procedure outlined in Section~\ref{sec:arc_para}, we calculate the arc length of each segment. Then, we compute the gradient for optimization using the numerical gradient calculation equation from Section~\ref{sec:gradient_calc}.

To ensure that the end orientation closely matches the target orientation, we also impose constraints on the orientation of the end effector of the continuum robot. We define the objective function guiding this constraint as the norm of the difference between the current orientation vector and the target orientation vector of the end effector. The calculation process for the gradient of each constraint is similar, allowing us to efficiently incorporate them into the optimization process.

%% file: sections/update_rules.tex
According to Section~\ref{sec:cost_function}, the overall cost function value can be calculated as:
$$U(\xi) = \alpha_1F_{obs}(\xi) + \alpha_2F_{len}(\xi) + \alpha_3F_{ori}(\xi)$$

Here, $\alpha_1$, $\alpha_2$, and $\alpha_3$ are changeable weights of the terms during optimization. We utilize the simulated annealing algorithm to adjust the weights during optimization. Additionally, we employ other techniques, such as projecting the update vector for obstacle avoidance onto a surface perpendicular to the plane generated by the current path point and the adjacent path points, to aid in convergence.

To perform gradient descent, we first approximate the cost function with a first-order Taylor expansion:

$$U(\xi) \approx U(\xi_0)+\Delta U(\xi_0)^T(\xi-\xi_0)$$

To minimize changes in the average 'acceleration' of each path point, which is defined to be proportional to the difference between its distances towards adjacent path points, we add a regularization term $\frac{\lambda}{2}(\xi - \xi_k)^T A (\xi - \xi_k)$, where $A$ is the acceleration matrix. Thus, the optimization problem becomes:

$$\xi =\mathop{\arg\min}\limits_{\xi}\left\{ U(\xi_0)+\Delta U(\xi_0)^T(\xi-\xi_0)+ \frac{\lambda}{2}||\xi-\xi_0||_A\right\};$$
$$\frac{\lambda}{2}||\xi-\xi_0||_A=\frac{\lambda}{2}(\xi-\xi_0)^TA(\xi-\xi_0)$$ 

Here, $\xi_0$ represents the current path points, and $\xi$ represents the updated path points.

$\lambda$ represents the regularization coefficient, serving as a trade-off between minimizing the cost function value and minimizing changes in the average acceleration of the existing trajectory, for better convergence. As a result, the gradient descent update rule can be written as:

$$\xi=\xi_0-\frac{1}{\lambda} A^{-1}\Delta U(\xi_0)$$

The update will continue until all constraints are satisfied or the maximum iteration number is reached.


    
      
      
      
      
      

%% file: sections/examples_n_results.tex
\subsection{Performance Comparison}
To provide a clear comparison of the time consumption, we conducted experiments using a 3D test environment modeled after the settings in a previous research paper \cite{chiang2021slinki}. In this environment, our IK solver was tested against others with 5 segments, a segment length range of 45~mm to 70~mm, and an acceptable end effector orientation error of 0.0175~rad, without obstacles. The start point's orientation was set to (0, 0, -1), and the four end orientations were (-0.7071, 0, -0.7071), (0, 0.2169, -0.9762), (0.5661, 0.2665, -0.7926), and (0, 0, 1) respectively.

The start point's position was set to (0, 0, 0), and the four end positions were (-60, 150, -230), (-10, 40, -260), (150, -90, -220), and (40, -60, -180) respectively, in mm. Each test was repeated 100 times for each setting to obtain the mean performance. Since our algorithm's nature guarantees the same solution, we aimed to achieve a more accurate estimate of time consumption.

\begin{figure}
    \centering
    \includegraphics[width=0.8\linewidth]{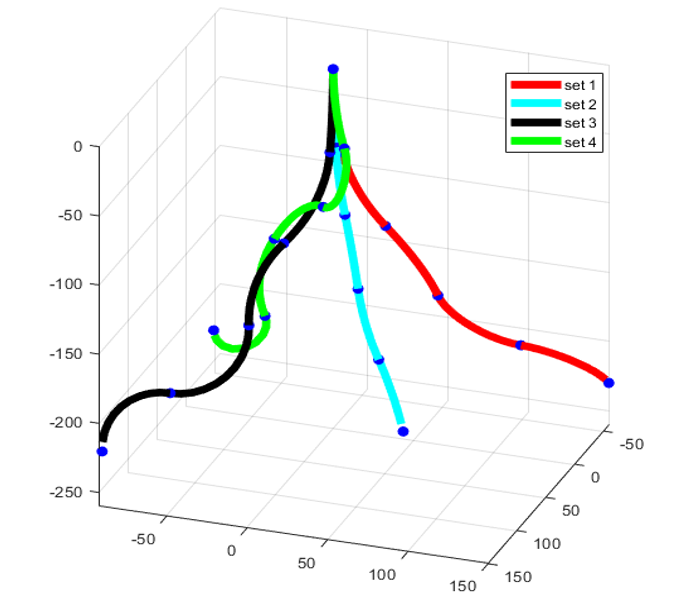}
    \caption{
    The result of the four settings by the IK solver introduced in this paper. The piece-wise arcs represent the configuration of the continuum robot calculated by our IK solver, under four different settings of end position and orientation.
}      
    \label{speed_comparison}
\end{figure}

The results are presented in Figure~\ref{speed_comparison}, and the data is summarized in Table~\ref{table:time_consumption}, focusing on the comparison with two specific sets as the previous paper did. Our algorithm is labeled as \textbf{Opti.path}.

Comparing the results, our algorithm demonstrates competitive time consumption, particularly excelling in 0 position error. However, being an optimization-based approach, its efficiency is influenced by the initial guess's proximity to the solution. For instance, in set 3, it took 15 iterations (15.9~ms) to find the solution, and even less if the initial guess was closer. Conversely, in set 4, where the solution was further from the initial guess, it took over 100 iterations, totaling 172~ms. This characteristic makes it suitable for IK during path tracking, as it can leverage previous results to quickly find nearby solutions, as we will demonstrate in the following section.

\begin{table*}
    \caption{IK Solutions Comparison Under Same Orientation and Pose Setting} 
	\centering 
	\begin{tabular}{c c c c c c} 
		\hline\hline 
		\multicolumn{6}{c}{Orientation: (-0.7071, 0, -0.7071); Position: (-60, 150, -230)}\\[1.0ex]
		\hline 
&time [ms]&position [mm]&pos-err [mm]&orientation&ori-err [rad]\\[1.0ex]
exFABRIKc&183&(-60,149,-231)&0.863&(-0.71,0,-0.71)&0\\[1.0ex]
Opti.DLS&112&(-62,152,-234)&4.52&(-0.4,0.77,-0.5)&0.85\\[1.0ex]
SLInKi&12.2&(-59,150,-230)&0.766&(-0.71,0,-0.71)&0\\[1.0ex]
\textbf{Opti.path}  &16.9&(-60,150,-230)&0&(-0.70,-0.01,-0.71)&0.014\\[1.0ex]
\hline 
\multicolumn{6}{c}{Orientation: (0, 0.2169, -0.9762); Position: (-10, 40, -260)}\\[1.0ex]
		\hline 
&time [ms]&position [mm]&pos-err [mm]&orientation&ori-err [rad]\\[1.0ex]
exFABRIKc&206&(-9.7,40,-261)&0.775&(0,0.22,-0.98)&0\\[1.0ex]
Opti.DLS&124&(-10,40,-260)&0&(0,0.22,-0.98)&0\\[1.0ex]
SLInKi&11.8&(-9.9,40,-259)&0.632&(0,0.22,-0.98)&0\\[1.0ex]
\textbf{Opti.path}  &2.8&(-10,40,-260)&0&(0.02,0.22,-0.98)&0.017\\[1.0ex]

    \hline 
	\end{tabular}
	\label{table:time_consumption} 
\end{table*}

\subsection{Obstacle Avoidance}
In this section, we create complex environments with multiple spherical obstacles, each positioned randomly with random radius. The goal is to guide a continuum robot through these obstacles without collision. We segment the path into 180 points, and the robot is tasked with tracking a path on the opposite side of the workspace through the obstacles.

To begin, we solve the IK from the start point to the path's initial point, establishing an initial state. We then commence the optimization process by adjusting the end point and recalculating the IK using our optimization-based solver. We set a tolerance for segment length deviation, limiting it to one-tenth of the original length. The base orientation remains constant, but there are no constraints on the end effector orientation due to the complex environment.

\begin{figure*}
    \centering
    \includegraphics[width=1.0\linewidth]{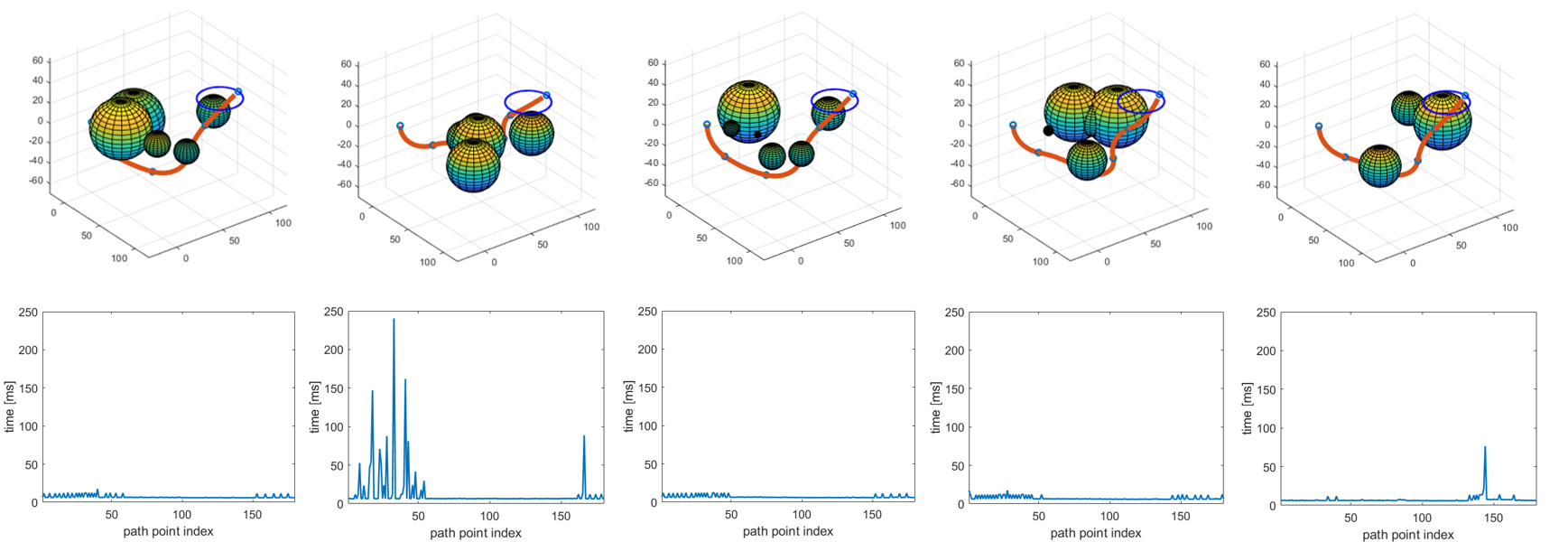}
    \caption{
    Upper figures depict the continuum robot successfully navigating various intricate environments filled with sphere obstacles. The red piece-wise arcs with blue end points represent the continuum robot configuration calculated by our IK solver, and the blue lines represent the path to be tracked. The lower plots provide a visual representation of the corresponding time consumption. The x-axis represents the path point index, while the y-axis indicates the time consumption for each path point, measured in milliseconds.
}      
    \label{obstacle_avoidance}
\end{figure*}

Each experiment is repeated 30 times for each setup, and the mean time consumption is calculated. The results, presented in Figure~\ref{obstacle_avoidance}, demonstrate an acceptable latency and the algorithm's ability to quickly reach local optimization during path tracking. Typically, solving IK for each path point takes between 6~ms and 20~ms. However, in narrow areas, this time can increase to over 100~ms, or even 200~ms.

Additionally, we designed other path configurations for the robot to track, showcasing different challenges. Due to page limitations, these configurations are not detailed here but can be viewed in the supplementary video.

\subsection{Time Consumption With Relate to Number of Segments}
\begin{figure}
    \centering
    \includegraphics[width=1.0\linewidth]{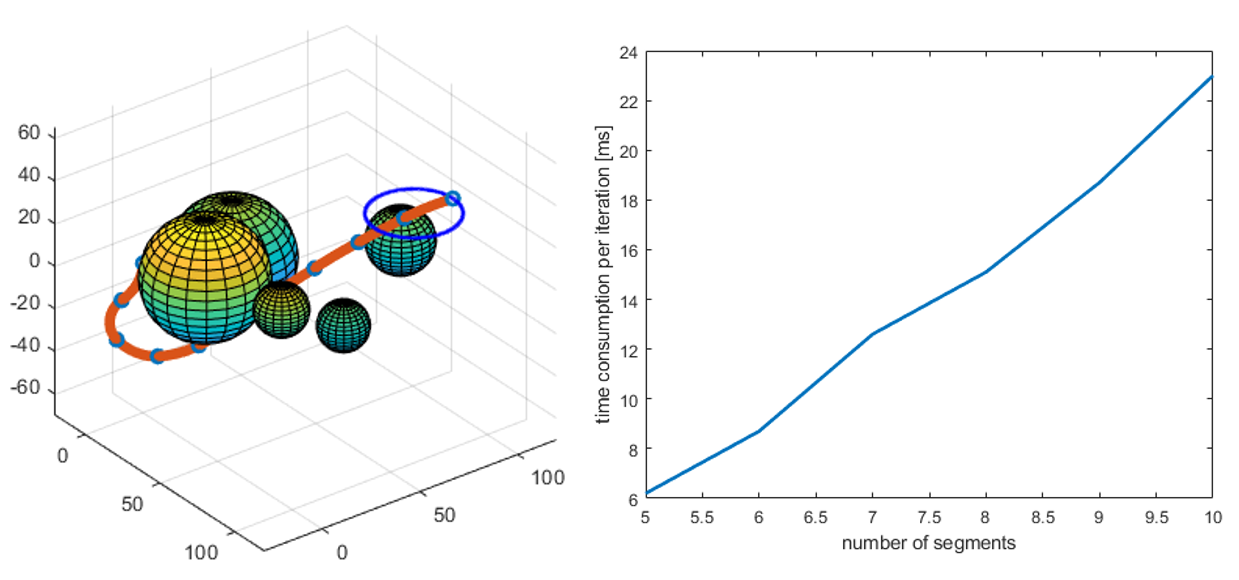}
    \caption{
    Left: A 10-segment continuum robot tracking a path while avoiding obstacles. Right: The relationship between the number of segments and the time consumption per iteration.
}      
    \label{seg_time}
\end{figure}

In order to demonstrate the potential for our robot to solve IK for hyper redundant continuum robot, we run the continuum robot in simulation to track path through an obstacle rich intricate environment, while the number of segment varies from 5 to 10. Each set of experiment runs 20 times and gather the mean value as our result. We measure the time consumption per iteration for them and the result is shown in Figure~\ref{seg_time}.
As shown in the figure, the time consumption doesn't increase too much as the number of segment increases, at least not exponentially increase.

In order to demonstrate the potential of our robot to solve IK for a hyper-redundant continuum robot, we conducted simulations where the robot tracked a path through a complex, obstacle-rich environment. We varied the number of segments from 5 to 10 and ran each experiment set 20 times each, averaging the results for analysis. The time consumption per iteration was measured, and the results are shown in Figure~\ref{seg_time}. Interestingly, the figure indicates that the time consumption does not increase significantly as the number of segments increases, at least not exponentially.

